\documentclass[conference]{IEEEtran}
\IEEEoverridecommandlockouts
\usepackage{cite}
\usepackage{amsmath,amssymb,amsfonts}
\usepackage{algorithmic}
\usepackage{graphicx}
\usepackage{textcomp}
\usepackage{xcolor}
\usepackage{multirow}
\usepackage{multicol}
\usepackage{adjustbox}
\usepackage{subfigure}
\def\BibTeX{{\rm B\kern-.05em{\sc i\kern-.025em b}\kern-.08em
    T\kern-.1667em\lower.7ex\hbox{E}\kern-.125emX}}
\begin{document}
\setlength\abovedisplayshortskip{0pt}
\setlength\belowdisplayshortskip{5pt}
\setlength\abovedisplayskip{5pt}
\setlength\belowdisplayskip{10pt}

\title{\huge``Stop Asian Hate!": Refining Detection of Anti-Asian Hate Speech during the COVID-19 Pandemic
}

\author{
\IEEEauthorblockN{Huy Nghiem, Fred Morstatter}
\IEEEauthorblockA{\textit{USC Information Sciences Institute} \\
Email:\{hnghiem, morstatt\} @usc.edu
}
}

\maketitle

\begin{abstract}
\textit{Content warning: This work displays examples of explicit and/or strongly offensive language.} Fueled by a surge of anti-Asian xenophobia and prejudice during the COVID-19 pandemic, many have taken to social media to express these negative sentiments. Identifying these posts is crucial for moderation and understanding the nature of hate in online spaces. In this paper, we create and annotate a corpus of tweets to explore anti-Asian hate speech with a finer level of granularity. Our analysis reveals that this emergent form of hate speech often eludes established approaches. To address this challenge, we develop a model and an accompanied efficient training regimen that incorporates agreement between annotators. Our approach produces up to 8.8\% improvement in macro F1 scores over a strong established baseline, indicating its effectiveness even in settings where consensus among annotators is low. We demonstrate that we are able to identify hate speech that is systematically missed by established hate speech detectors.   
\end{abstract}

\begin{IEEEkeywords}
covid, pandemic, social media, hate speech, deep learning
\end{IEEEkeywords}

\section{Introduction}
Since its first reported incident at the end of 2019, the coronavirus (COVID-19) pandemic has had tremendous global economic impact and led to an unprecedented disruption in the daily lives of billions  \cite{wu2020outbreak}. Historically, outbreaks of this nature have been associated with xenophobia and othering of certain groups \cite{le2020anti}. That the virus first emerged from Wuhan, China, along with allegations of its chiropteran origins  have only served to fuel xenophobic attitude against Asians in Western countries \cite{joubin2020anti, kwan2020understanding}. In the United States (US), Asians and Asian Americans not only have to contend with the risk of infection, but also the target of prejudice and discrimination against their ethnicity \cite{reny2022xenophobia}. According to reports by the Federal Bureau of Investigations (FBI), hate crimes against Asians Americans increased by 77\% from 2019 to 2020 \cite{kapadia2022violence}. Asian groups have been reported to experience higher level of mental heath disorders since the onset of the pandemic than other races \cite{wu2021anti}. 

Anti-Asian xenophobia not only manifests in physical settings, but also on social media.  Tahmasbi et al. \cite{tahmasbi2021go} also discovered a significant rise of old and new Sinophobic slurs on Twitter induced by the coronavirus pandemic. A study on Twitter \cite{hswen2021association} finds that tweets that contain the hashtag \textit{``\#chinesevirus"} are more than twice as likely than those with \textit{``\#covid19"} to contain anti-Asian sentiments. That even prominent political figures adopted these terms further encouraged their usage and incited discriminatory associations \cite{budhwani2020creating}. Zhong, Huang and Liu \cite{zhong2021mental} found an association between an increased mental health toll and social media usage of residents of Wuhan, China -- the first epicenter of the outbreak. 

Detection of hate speech has received increased attention due to the prevalence in social media.  In spite of the new anti-Asian varieties of hate speech born from this pandemic, there are comparatively much fewer Asian-focused literature in this area \cite{ayo2020machine, misra2020psychological}. In this paper, we attempt to address this gap in research. Our contributions to hate speech literature are as follows:
\begin{itemize}
    \item Creation of an Asian-focused, COVID-oriented Twitter corpus with annotation on multiple dimensions of abusive and hate speech.
    \item Proposal of deep learning models that leverage individual annotations to enhance text classification amidst low agreement. 
    \item Analysis of efficient multitask training regimens to harness the benefits of these architectures.
\end{itemize}


\section{Related Work}
\subsection{Hate Speech Detection}
Hate speech detection is complex and challenging. A set of unifying standard definitions is yet to exist, leaving much room for subjectivity for discerning what constitutes hate and abusive speech. There are also overlapping -- and at times competing -- definitions from related works on offensive, toxic, hostile or prejudice speech that present added difficulty for generalizability \cite{macavaney2019hate}. In this work, we use the following definitions put forth by \cite{founta2018large}: 
\begin{itemize}
     \item \textbf{Abusive Speech} : ``Any strongly impolite, rude or hurtful language using profanity, that can show a debasement of someone or something, or show intense emotion.''
     
    \item \textbf{Hate Speech:} ``Language used to express hatred towards a targeted individual or group, or is intended to be derogatory, to humiliate, or to insult the members of the group, on the basis of attributes such as race, religion, ethnic origin, sexual orientation, disability, or gender.''

\end{itemize}

Literature on hate speech often focuses on manifestations of racism, sexism and discrimination against minority groups \cite{fortuna2018survey}. Models on hate speech detection have also grown in diversity and complexity, ranging from  Logistic Regression and Bayesian Networks to Genetic Algorithm and Deep Neural Networks, from singular to ensemble classifiers \cite{ayo2020machine, fortuna2018survey}.

\subsection{Annotation Strategies and Label Usage}
Hate speech and related textual corpora are often constructed by filtering from a larger source (usually some social media platforms) using predetermined keywords \cite{poletto2021resources, macavaney2019hate}. Twitter is the most common platform where researchers extract their original content \cite{vidgen2020directions}.  Labels for these corpora are often solicited from experts and trained workers, who are often more costly due to their training, or crowdsourcing, which is relatively cheaper and easier, or a mixture of both of these approaches \cite{vidgen2020directions}. Earlier efforts on hate speech often focused on binary annotation schemes (e.g.: \textit{hate} vs. \textit{not}) whereas more recent works explored multi-level and/or multi-class schemes with higher levels of granularity \cite{poletto2021resources, vidgen2020directions}.  

Due to the inherent ambiguity of language and the subjectivity of hate speech, disagreement among annotators for the same inputs occurs naturally. Researchers have noted that reported inter-rater agreement metrics \cite{poletto2021resources, vidgen2019challenges} , such as Cohen’s $\kappa$, Fleiss’ $\kappa$, Krippendorf’s $\alpha$, are often low. While earlier works often treated samples with low annotation agreement as noise, recent trends have found that such disagreements exemplify real difficult scenarios that should be represented rather than forcibly adjudicated or discarded \cite{leonardelli2021agreeing}.

Majority voting and averaging are typical approaches to derive the final label when disagreement among annotators is present. However, emergent research \cite{leonardelli2021agreeing, larimore2021reconsidering, zhou-etal-2021-challenges} has pointed out the shortcomings of using only these hard labels, such as biasing against minority voices and yielding internally inconsistent labels, which may in turn negatively impact models' performance. Fornaciari et al. \cite{fornaciari2021beyond} proposed a multitask model that incorporated probability distributions over the annotator labels to more accurately predict gold labels in Part-of-Speech tagging and morphological stemming tasks. For hate speech, Davani, Diaz and Prabhakaran \cite{davani2022dealing} demonstrated that predicting each individual annotator's label separately using a shared architecture also resulted in improved performance over predicting only aggregate labels. 


\section{Data Methodology}

\subsection{Data Acquisition}
We construct our corpus from the \textit{Public Coronavirus Twitter Dataset} \cite{chen2020tracking}. Starting on January 28, 2020, the collectors leveraged Twitter's API to collect tweets that contain relevant keywords. This repository is updated on a weekly basis\footnote{Available at https://github.com/echen102/COVID-19-TweetIDs}, and contains over 150 million Tweet IDs (proprietary unique identifier of each tweet) as of December 2021.  As hate speech occurrences are proportionately rare on Twitter, we utilize a boosting approach proposed by \cite{founta2018large} that involves sampling using targeted keywords. First, we collect a set of anti-Asian phrases that include both existing (\textit{e.g.: ch*nk, g*ok}) and emergent, COVID-related (\textit{e.g.: china lied people die, chinavirus}) from \textit{Hatebase.org} and relevant literature \cite{tahmasbi2021go, he2021racism, vidgen2020detecting}. We also collect a similar set for anti-Black phrases out of an interest in interracial interaction during COVID. After inspection, the final collection contains  49 anti-Asian phrases, 56 anti-Black phrases (detailed in the Appendix. To the best of our knowledge, this list -- while non-exhaustive -- consists of the most representative phrases targeted towards Asian and Black demographics in hate speech literature. 

Each candidate tweet is tentatively categorized as either \textit{anti-Asian} or  \textit{anti-Black} if it contains a phrase of the corresponding set, \textit{Interracial} if phrases from both sets are present, and \textit{Normal} if none applies. We restrict our sampling pool to only original tweets written in English from the general corpus. To compensate for the natural imbalance of hateful tweets (the positive classes), we sequentially select at random a tweet from each of the 4 classes until we reach 500 tweets for each month from July to December 2020. The final corpus has 3000 tweets, which are selected for annotation. 

\subsection{Data Annotation}
\begin{figure}[htbp]
\centerline{\includegraphics[width=\columnwidth]{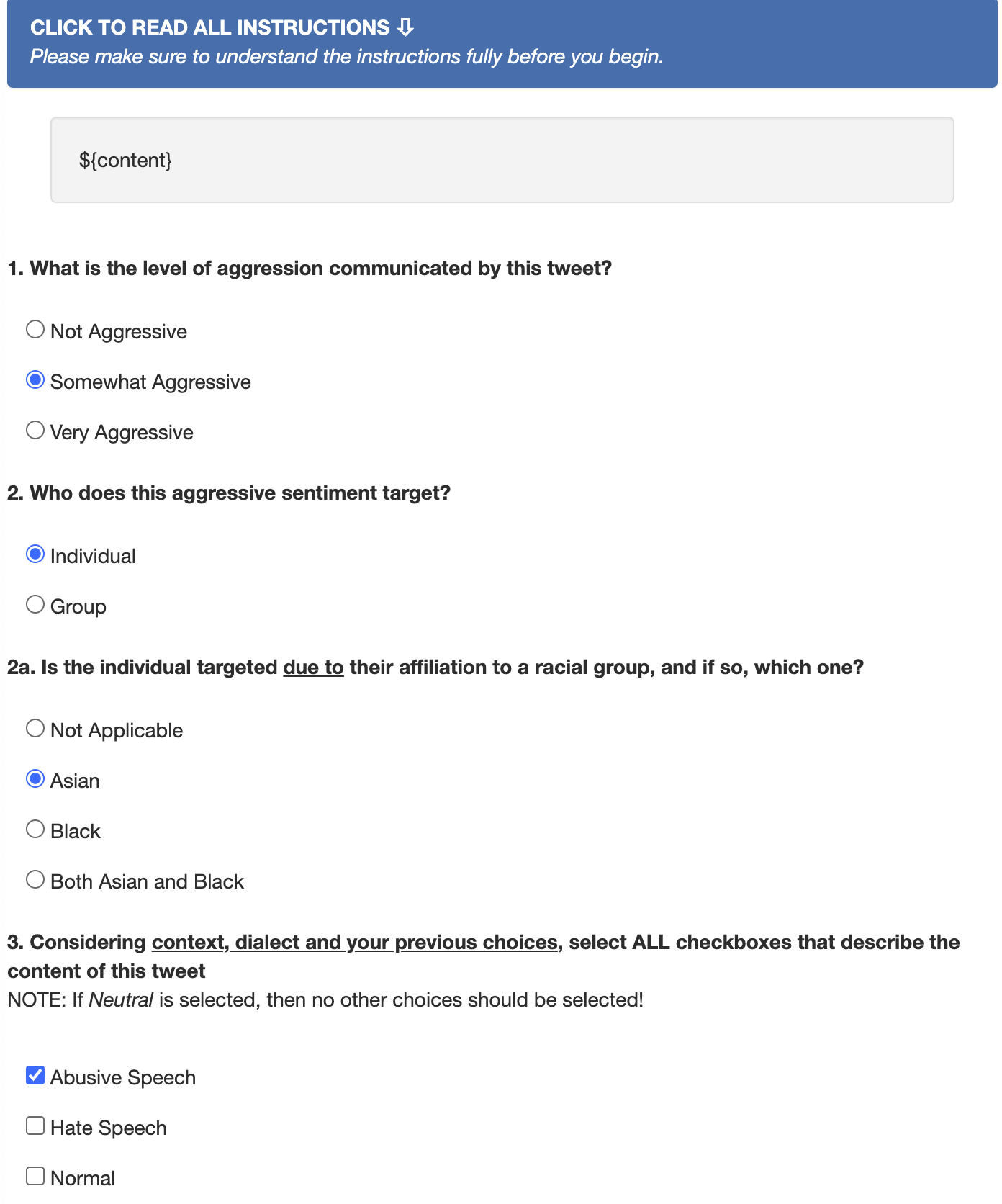}}
\caption{Template for AMT Annotation.}
\label{fig:template}
\end{figure}

Fig. \ref{fig:template} illustrates the general template for annotation. At the top is a collapsible instructional pane, followed by the tweet's content and then prompts to identify the following attributes:
\begin{itemize}
    \item \textbf{(Level of) Aggression}: \textit{Not Aggressive, Somewhat Aggressive, Very Aggressive}
    \item \textbf{Target}: \textit{Not Applicable (N/A), anti-Asian, anti-Black, Both anti-Asian and anti-Black}
    \item \textbf{(Speech) Type}: \textit{Abusive, Hate, Normal}
\end{itemize} 
While \textit{Type} (or its equivalent) is a standard attribute appearing in other works on hate speech, we include the other 2 attributes in an effort to examine the direction and strength of the expressed negative sentiments. 

Inspired by \cite{sap2019risk}, we prime annotators at the beginning of the instructions to consider contextual clues: \textit{``Note on Dialect and Context:
Some words may be considered offensive generally, but not in specific contexts, especially when used in certain dialects or by minority groups.
Please consider the context of the tweet while selecting the options below."}
We then provide definitions and an example for each option. In an effort to enhance consistency in the field, we decide to use verbatim the definitions as provided by \cite{founta2018large} and \cite{chatzakou2017mean} for the questions on \textit{Type} and \textit{Level of Aggression}, respectively.Following examples in \cite{assimakopoulos2020annotating, zampieri2019predicting}, we present our questions in a hierarchical fashion to direct annotators' consideration  on different aspects laid out in the preceding definitions. We first inquire annotators about the \textit{Level of Aggression} communicated by the tweet. Only if a choice other than \textit{Not Aggressive} is selected then subsequent questions are displayed. Otherwise, subsequent questions would return default neutral options upon submission. 

The task of identifying the target of aggression is split into two parts: first to identify whether the target is an \textit{Individual} or \textit{Group}. If the former, we inquire about the target's racial affiliation as shown in Fig.\ref{fig:template}. If the latter, we urge the annotators to consider context and dialect of the tweet again as they identify the target's race . Finally, we ask annotators to identify the tweet's \textit{Type}. This question allows selection of both \textit{Abusive} and \textit{Hate} options, and only \textit{Normal} otherwise. We present this question last to allow annotators the opportunity to deliberate based on their preceding options. 

\subsection{Crowdsourcing} 
We solicited the templates to workers on Amazon Mechanical Turk (AMT), a popular crowdsourcing platform for this type of task. A Human Intelligence Task (HIT) consists of a single set of responses on a template attached to a tweet drawn from our corpus. Each HIT requires 3 sets of  annotation from workers to be considered completed. We required workers to possess the following qualifications: HITs approval rate at least 95\%, locations in either Canada or the U.S, and consenting to viewing adult themes. In addition, we asked annotators to complete a pre-screen survey on their demographic backgrounds, including \textit{Age Group, Gender, Ethnicity} and \textit{Level of Education} . Workers received \$.06 for successful submission of a HIT and was capped at approximately 300 HITs to ensure diversity of opinions. The discussion of our annotation's background is out of the scope of this paper. However, this information will be made available along with our data for interested researchers. 

\subsection{Annotation Results and Analysis}
Table \ref{tab:annot_results} reports the results of our annotation tasks. For inter-annotator agreement, we observe Fleiss' $\kappa$ values in the range of 0.41 to 0.49. We use majority voting to derive the singular labels for each tweet's respective attribute, whose aggregates are reported in $Count$. When ties happen for \textit{Aggression}, the final label is defaulted to \textit{Somewhat Aggressive}. For \textit{Type}, $Abusive$ is the default choice when votes tie. We observe that the number of non-neutral labels all belong to the minority of classes. 

\begin{table}[]
\caption{Annotation Results. \textit{(A.Count: Adjusted Count)}}
\begin{tabular}{|l|l|c|c|c|}
\hline
\multicolumn{1}{|c|}{\textbf{Attribute}} & \multicolumn{1}{p{2.7cm}|}{\hfil\textbf{Category}} & \textbf{$\kappa$} & \multicolumn{1}{|p{1cm}|}{\hfil\textbf{Count }} & \multicolumn{1}{|c|}{\hfil \textbf{A. Count}} \\ \hline
\multirow{3}{*}{\textbf{Aggression}} & Not Aggressive       & \multirow{3}{*}{0.42} & 2187 & 2187 \\ \cline{2-2} \cline{4-5} 
                                     & Somewhat Aggressive  &                       & 725  & 691  \\ \cline{2-2} \cline{4-5} 
                                     & Very Aggressive      &                       & 88   & 81   \\ \hline
\multirow{4}{*}{\textbf{Target}}     & Not Applicable (N/A) & \multirow{4}{*}{0.49} & 2505 & 2505 \\ \cline{2-2} \cline{4-5} 
                                     & anti-Asian           &                       & 454  & 454  \\ \cline{2-2} \cline{4-5} 
                                     & anti-Black           &                       & 35   & N/A    \\ \cline{2-2} \cline{4-5} 
                                     & Both                 &                       & 6    & N/A    \\ \hline
\multirow{3}{*}{\textbf{Type}}       & Normal               & \multirow{3}{*}{0.41} & 2379 & 2373 \\ \cline{2-2} \cline{4-5} 
                                     & Abusive              &                       & 382  & 354  \\ \cline{2-2} \cline{4-5} 
                                     & Hate                 &                       & 239  & 232  \\ \hline
\end{tabular}
\label{tab:annot_results}
\end{table}

\noindent{\textbf{Perspective API}} We use \texttt{Perspective API} \footnote{Details on the following descriptions can be found at https://developers.perspectiveapi.com}, developed by the Google Counter Abusive Technology and Jigsaw team, to investigate this commercial tool's behaviors on our dataset. Trained on millions of comments from various forums, the API produces a probability score for certain attributes of abusive language, which represents the fraction of raters who find the corresponding attribute present in the input \cite{perspective}. 

\begin{figure}[htbp]
    \centering
    \includegraphics[width=0.325\columnwidth]{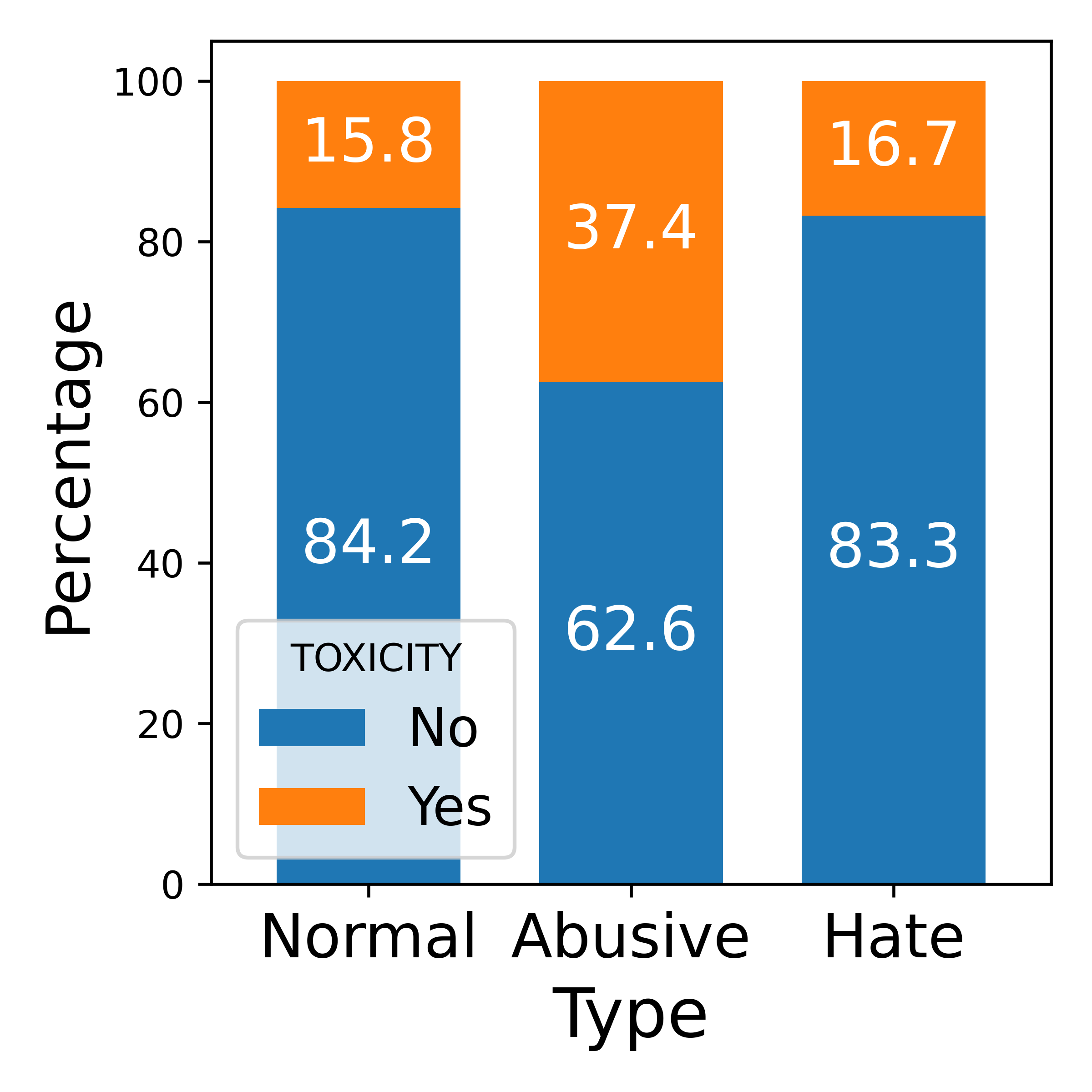}
    \includegraphics[width=0.325\columnwidth]{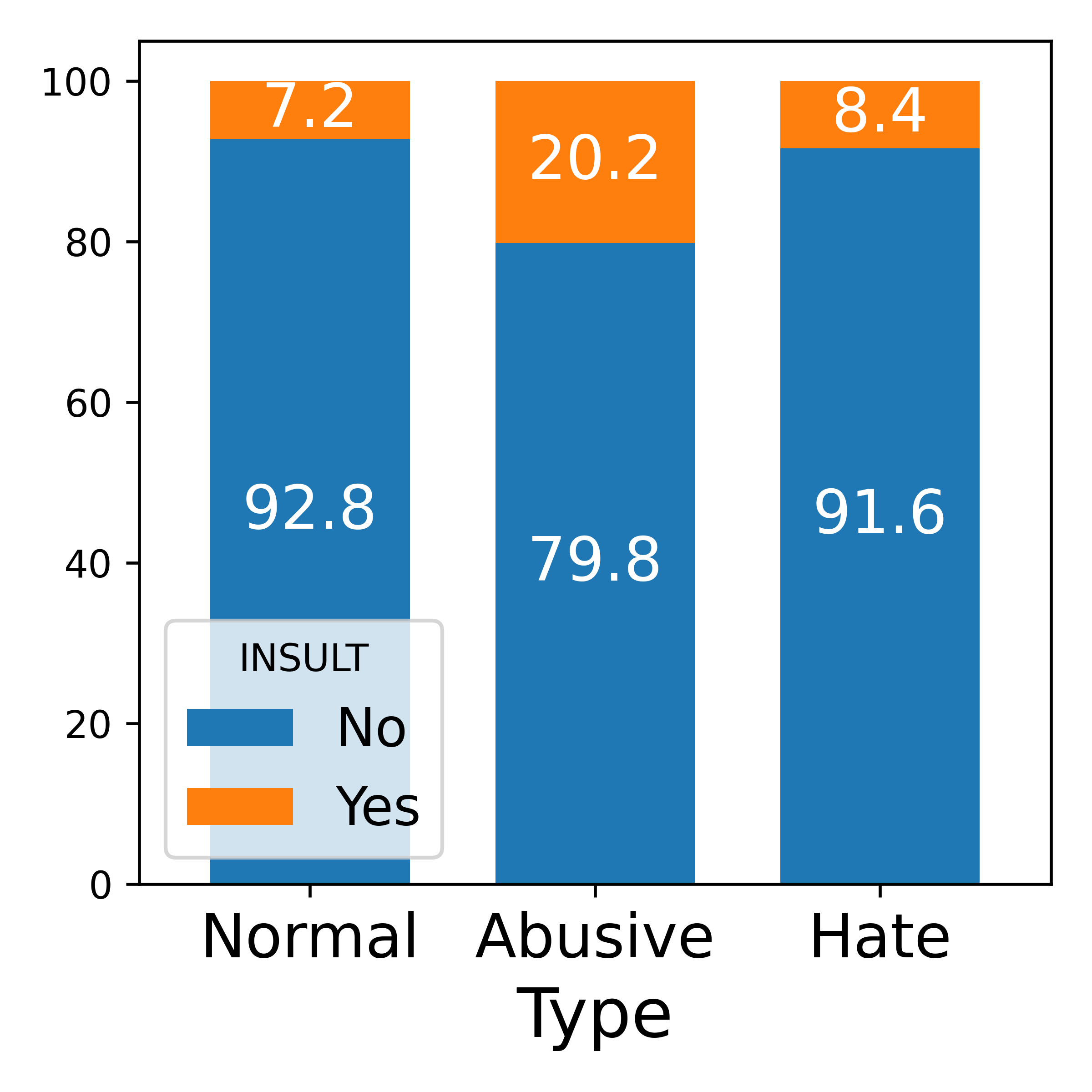}
    \includegraphics[width=0.325\columnwidth]{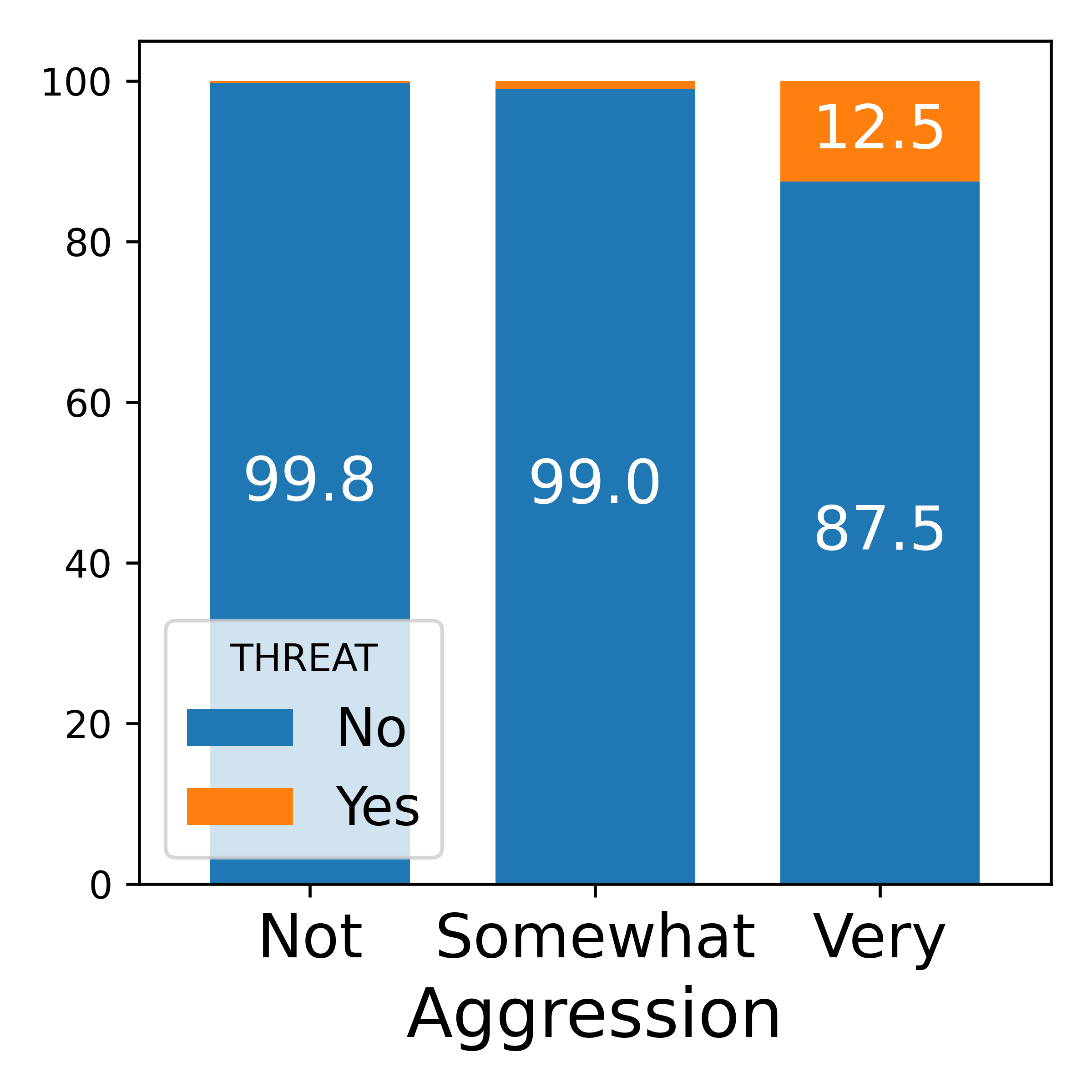}
    \caption{Breakdown of selected Perspective's API scores for our attributes.}
    \label{fig:pers}
\end{figure}

We request the scores for the attributes \textit{Toxicity, Insult} and \textit{Threat} from the API based on their definitions and relevance to this study. Using the cutoff probability 0.5 (\texttt{Perspective} suggests 0.7 \cite{perspective}), tweets with score greater than this threshold are assigned a value of \textit{Yes} for that attribute, otherwise \textit{No}. The stacked bar charts in Fig. \ref{fig:pers} shows the breakdown of percentages of \texttt{Perspective}'s predictions for our annotated \textit{Type} and \textit{Aggression} attributes. The discrepancy is apparent: \texttt{Perspective} significantly predicts our non-neutral categories to \textit{not} contain their negative attributes. More interestingly, tweets that our annotators considered \textit{Abusive} are more frequently considered positive for \textit{Toxicity} and \textit{Insult} by the API compared to their \textit{Hate} counterpart. Furthermore, $\sim 16$\% and 7\% of our \textit{Normal} tweets are considered toxic. \texttt{Perspective} also only label about 12\% of \textit{Very Aggressive} tweets to be \textit{Threat} while only 1\% of \textit{Somewhat Aggressive} is considered so. Though sophisticated and heavily trained, \texttt{Perspective} may not yet capture COVID-related societal and linguistic concepts.

\noindent{\textbf{Towards Asian Focus}} Forty-one tweets whose  \textit{Target} are \textit{anti-Black} and \textit{Both} account for only $\sim$ 1.4\% of our corpus. We decide to remove those tweets from our final pool to focus on Asian groups in our subsequent analyses. For the remaining tweets, we convert the leftover individual labels for the \textit{Target} attribute from \textit{anti-Black} to \textit{Not Applicable} and from \textit{Both} to \textit{anti-Asian}. Final results are shown in the \textit{Adjusted Count (A. Count)} column in Table \ref{tab:annot_results}.

\begin{figure}
    \centering
    \includegraphics[width=0.325\columnwidth]{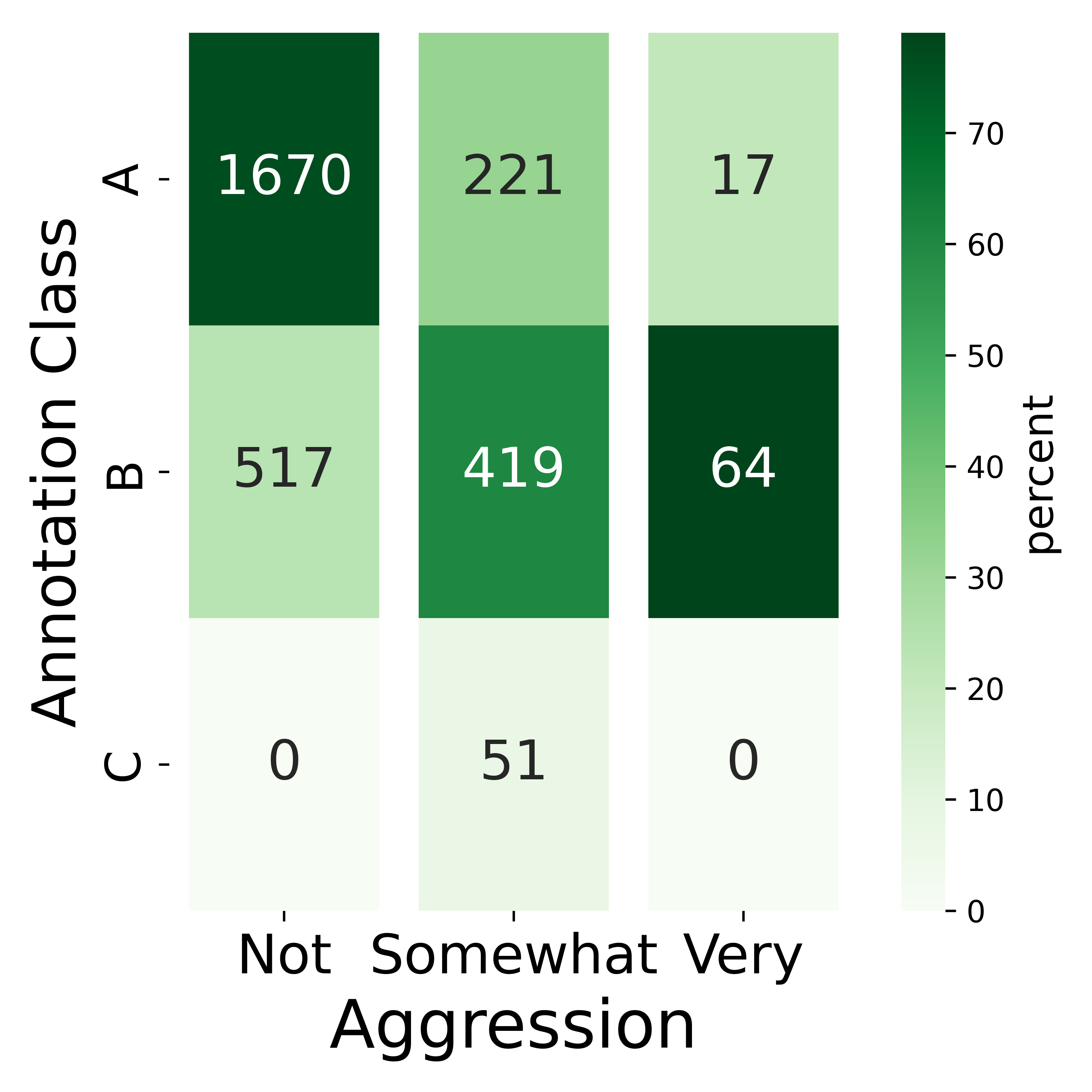}
    \includegraphics[width=0.325\columnwidth]{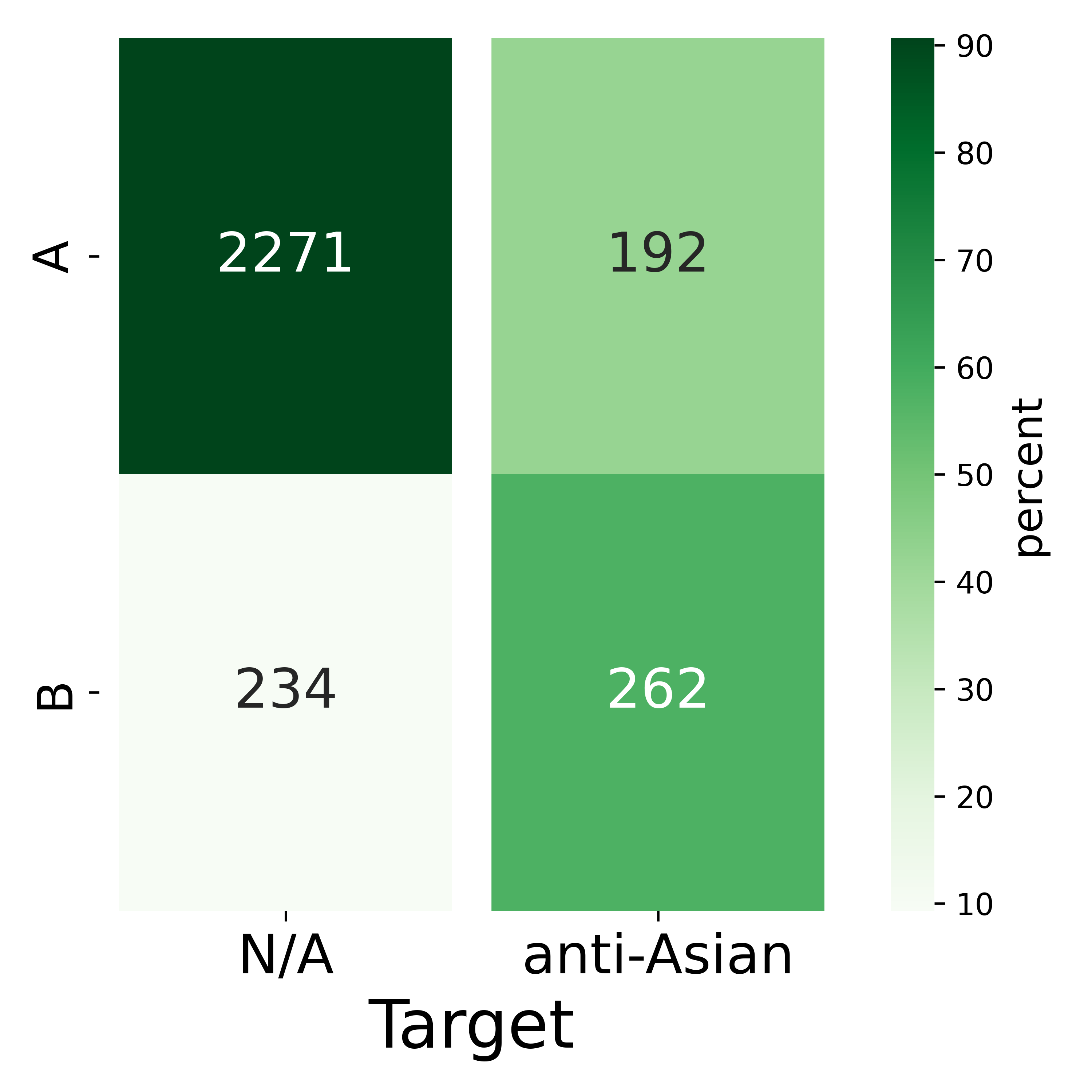}
    \includegraphics[width=0.325\columnwidth]{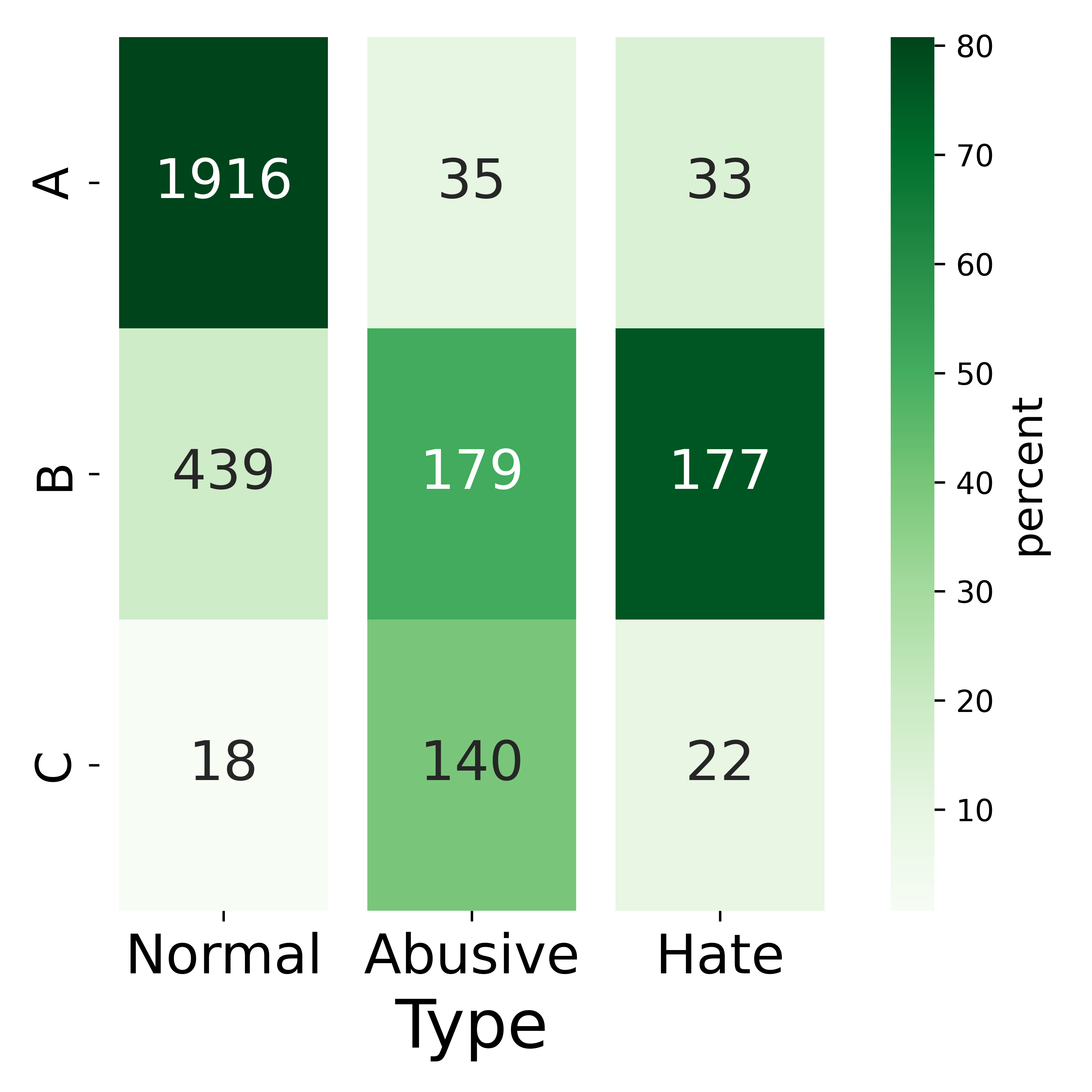}
    \caption{Cross-tabulations between our Attribute vs. Annotation Class. Color gradient based on column-wise percentages. }
    \label{fig:annot_class}
\end{figure}

\noindent{\textbf{Annotation Class}} For each attribute, each of the remaining 2959 tweets is assigned a value for \textit{Annotation Class} based on the level of agreement between annotators: \textit{A} if all annotators unanimously agree, \textit{B} if annotators differ on exactly 2 categories, and \textit{C} if each of the attribute's available categories is selected by at least 1 annotator. Fig. \ref{fig:annot_class} displays the contingency tables between our attributes and their corresponding \textit{Annotation Classes}. The background color gradient correlates to each cell's column-wise percentage. 

We observe that \textit{Target} only contains 2 \textit{Annotation Classes}. For the \textit{Not Aggressive} and \textit{Very Aggressive} categories, \textit{Class A} and \textit{B} dominate respectively, indicating strong general agreement. On the other hand, \textit{Somewhat Aggressive} sees 51 instances (7.3\% of this category) where annotators could not reach consensus (\textit{Class C}) and a majority of this category (61\%) falls within \textit{Class B}. More pronounced levels of disagreement is observed for \textit{Type} attribute. While \textit{Class C} instances appear in both \textit{Neither} and \textit{Hate} categories, this class slightly trails behind \textit{Class B} (40\% vs. 51\%) whereas \textit{Class} A occupies only 10\% of the \textit{Abusive} category. In addition to the class imbalance of final aggregate labels, these statistics reveal another dimension of implicit imbalance of annotator agreement. 

\noindent{\textbf{Comparison to other Datasets}} Among recent works on COVID-related hate speech, He et al. \cite{he2021racism} employed 2 undergraduate annotators to label 3255 tweets on anti-Asian hate and counter-hate speech under the authors' supervision. Vidgen et al.  \cite{vidgen2020detecting} used a team of experienced annotators from United Kingdom, Europe and South America to obtain annotations on \textit{Hostility, Criticism, Discussion and Counter speech} of East-Asian entities, with adjudication from PhD-level experts. An et al. \cite{an2021predicting} focused on characterizing anti-Asian hateful Twitter users. Complimenting these pioneering research our dataset offers more levels of granularity using crowdsourced annotation instead of experts. We believe that the varying levels of agreements demonstrated in the preceding section reflect the subtle diversity of opinions in the discussion of COVID-related hate speech. Thus, we strive to incorporate this information in our subsequent classification tasks.

\section{Model Development}
In this section, we develop various models to classify the tweets according to our attributes. From this point, we refer to the classification of the attributes \textit{Aggression}, \textit{Target} and \textit{Type} as \textbf{Task 1, 2} and \textbf{3}, respectively. We further convert the annotations for each tweets into probability distributions, or \textbf{soft labels}, while the \textbf{silver label} is derived from majority voting. For instance, a tweet with a 3-way split among \textit{Normal, Abusive} and \textit{Hate} for the \textit{Type} attribute would have the soft label $(0.\bar{3}, 0.\bar{3}, 0.\bar{3})$ and the silver label \textit{Abusive}. The soft labels' indices are arranged in the same order as their corresponding attribute introduced in Section III. 

\subsection{Tweet Pre-processing}
To be consistent with the content seen by AMT workers, we remove all non-ASCII character from the tweets and convert them to lower case. We further remove all retweet identifiers (\textit{`rt'}), and substitute hyperlinks with the $ \langle url \rangle$ token. All patterns of emoticons and emojis are also removed. Mentions of other usernames are replaced with the $\langle user \rangle$ tokens. Groups of repetitive patterns are replaced with a single representative (e.g.: \textit{``a a a"} becomes \textit{``a"}). Since hashtags may express relevant sentiments to our tasks, we process them with a specialized approach. We first remove the `\textit{\#}' sign. If a hashtag is in our list of hate phrases or \cite{chen2020tracking}'s specified keywords, we keep them as is. For other hashtags, we segment them into separate tokens using the Ekphrasis\footnote{Available at https://github.com/cbaziotis/ekphrasis} Python library.

\subsection{Architectures}
Architectures based on BERT (Bidirectional Encoder Representations from Transformers) have proven particularly versatile in sentiment-related classification tasks \cite{fortuna2018survey}. In our work, we employ RoBERTa (Robustly Optimized BERT Approach) as the basis of our models. In contrast to BERT, RoBERTA is pre-trained with only the Masked Language Modeling (MLM) objective on large English corpora, allowing the model to learn contextual word representations that are highly leverageable for downstream tasks. 

\begin{figure}[htbp]
\subfigure[General model architecture]{
\includegraphics[width=\columnwidth]{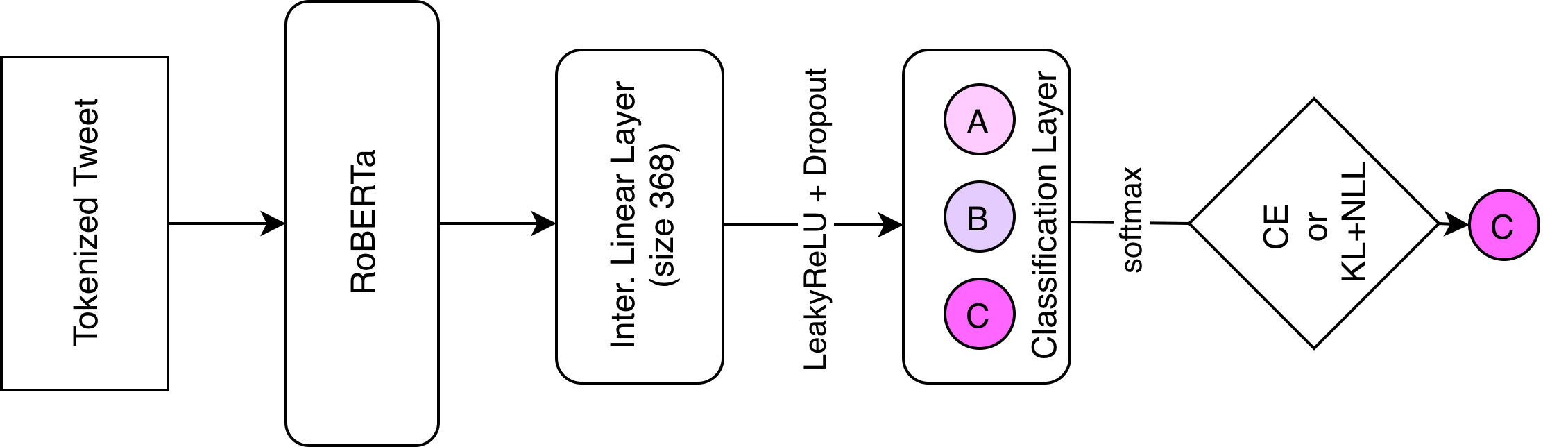}
\label{fig:modelarch_a}
}
\subfigure[Multitask training. Activation functions (not shown) similar to above.]{
\centerline{\includegraphics[width=\columnwidth]{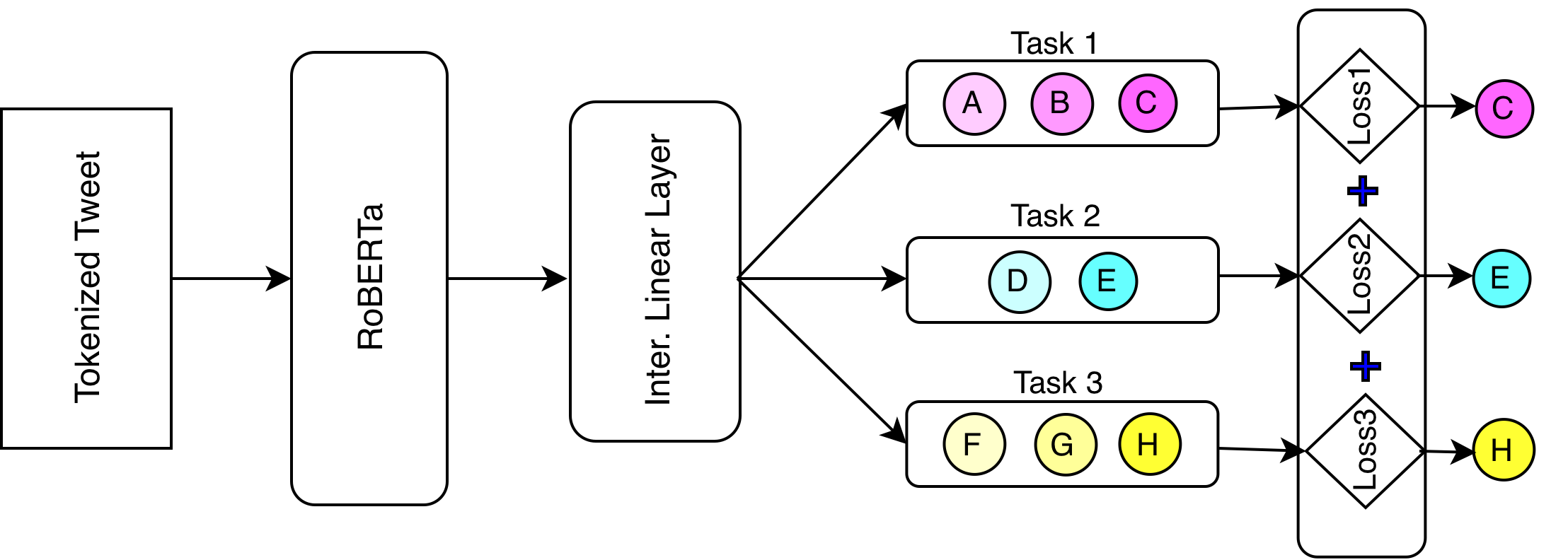}}
\label{fig:modelarch_b}
}
\caption{Diagrams of model architectures}
\label{fig:modelarch}
\end{figure}

To accommodate computational constrains, we choose RoBERTa\textsubscript{base} from the HuggingFace library, which contains 12 layers of transformers blocks, 12 attention heads and approximately 125 million trainable parameters.
Fig. \ref{fig:modelarch_a} illustrates the following general architecture. Processed tweets are tokenized by the RoBERTaTokenizer, then fed in mini-batches as inputs to to the RoBERTa\textsubscript{base} model. We keep the last-layer hidden state (size 768) of the sequence's classification token [CLS] and connect it to an intermediate linear layer of size 364. We apply the \textit{LeakyReLU} activation function and \textit{DropOut} to this layer's outputs before connecting them to the final linear classification layers, whose sizes correspond to the number of categories in each task. A \textit{Softmax} function is applied to this layer's outputs to normalize logits into final predicted probabilities. We distinguish the following models based on their objective functions.

\textbf{RoBERTa\_CE} This model seeks to minimize the classic Cross Entropy (CE) loss function. Equation \eqref{eqce} describes the loss function for each mini batch, where N denotes the number of samples in each batch, $p(\hat{y}|x)$ the model's predicted probabilities, and $y\textsubscript{silver}$ the singular silver label as previously described.

\begin{equation} 
\mathcal{L} = \frac{1}{N}\sum_{i=1}^{N}CE(\;p(\hat{y}|x)\,||\, y\textsubscript{silver}\;)
\label{eqce}
\end{equation}

\textbf{RoBERTa\_KLNLL} In contrast, this model simultaneously minimizes a combination of the Kullback–Leibler Divergence (KL) and Negative Log-likelihood (NLL) loss functions. The KL loss provides a measure of how one probability distribution is different from another while the NLL function calculates the negative logarithm of the model's predicted probabilities. It is important to note that we use the PyTorch implementation of NLL loss, which only considers the predicted probability of the correct class, i.e.: $ NLL = -\log{p(\hat{y}\textsubscript{correct})}$. In \eqref{eqklnll}, $\alpha$ and $\beta$ are scalars that control the contribution of each function to the final loss value. We also apply a \textit{Softmax} function to the raw soft label distributions, denoted by $p\textsubscript{softmax}(y)$, to be the target of the KL loss. This objective function is designed to incorporate soft labels, which contains information about agreement among annotators, and silver labels, which are derived from soft labels.

\begin{equation} 
\mathcal{L} = \frac{1}{N}\sum_{i=1}^{N}\Big{[} \alpha *  KL(\;p(\hat{y}|x)\,||\,p\textsubscript{softmax}(y) + \beta * NLL(\;p(\hat{y}\textsubscript{correct}))\Big{]}
\label{eqklnll}
\end{equation}

\noindent{\textbf{Class Weights}} To counteract class imbalance, we incorporate class weights into our loss functions. For each task, each sample is attributed a weight corresponding to its category as described in \eqref{eqweight}, where $S$ is the total number of observations in the training set, $C$ the number of categories for that task (attribute) and $s\textsubscript{c}$ the number of samples in that category $c$. Weights are applied to the corresponding samples in the calculation of the CE and NLL loss functions above.

\begin{equation} 
w\textsubscript{c} = \frac{S}{C * s\textsubscript{c}}
\label{eqweight}
\end{equation}

\subsection{Training Regiment}
\noindent{\textbf{Evaluation Metrics}} Precision, Recall and F1 scores are chosen as performance metrics due to to the  multi-categorical and imbalance nature of our tasks. In \eqref{eqprf}, TP stands for \textit{True Positives}, FP \textit{False Positives} and FN \textit{False Negatives}. Precision (P) gives the proportion of correct instances (TP) instances out of all samples that are predicted to belong to a category. On the other hand, Recall (R) gives the proportion of correctly predicted instances out of all samples that belong to the silver label. F1 score is the harmonic mean of Precision and Recall. Our models are evaluated on \textbf{macro} metrics, the numerical means of Precision, Recall and F1 respectively, over all categories of the corresponding task.

\begin{equation}
P=\frac{TP}{TP + FP}\;\;\; R=\frac{TP}{TP + FN} \;\;\;F1=2*\frac{P*R}{P+R}
\label{eqprf}
\end{equation}

\noindent{\textbf{Training Details}} We train our models using Google Collab Pro with GPU Tesla P100 with 16GB of RAM. The data is split using approximately 80:20 ratio, with 2400 samples in the training set and 559 in the test set. Models are optimized using the AdamW optimizer. \textit{Dropout}'s probability is set to 0.2. We tune hyperparameters using 4-fold cross-validation on the training set to select values that yield the best macro F1  metrics. The following sets are candidates for tuning with chosen values in bold: learning rate $\in$ \{1e-5, 2e-5, 3e-5\}, batch size $\in$ \{10, 20, 30\}, maximum length of sequence $\in$ \{150, 200, 250, 300\}, number of epochs $\in$ \{2, 3, 4, 5\}, $\alpha$ and $\beta \in$ \{0.2, 0.5, 0.8, 1\}  . Specific values are reported in the sections below. Once determined, the models are trained using these sets on the entire train data and evaluated on the test data. 

\subsubsection{\textbf{Single Task Training}} We perform experiments to examine whether the best performance is achieved when the tasks are trained simultaneously or individually. For this regiment, the models are trained independently for each of the 3 tasks. Loss functions are evaluated with respect to the soft and silver labels of the corresponding tasks. Table \ref{tabparam} reports the specific values for the chosen hyperparameters for each model.

\begin{table}[htbp]
\caption{selected hyperparameters for single task training \\ (for model names, R is short for Roberta)}
\begin{center}
\begin{tabular}{|l|c|c|c|c|c|c|c|}
\hline
\multicolumn{1}{|c|}{\textbf{\centering Model}}&
  \textbf{Task} &
  \multicolumn{1}{|p{0.5cm}|}{\textbf{\centering Learn. \\ Rate}} &
  \multicolumn{1}{|p{0.55cm}|}{\textbf{\centering Batch \\ Size}} &
  \multicolumn{1}{|p{0.5cm}|}{\textbf{\centering Max \\ Len.}} &
  \multicolumn{1}{|p{1cm}|}{\textbf{\centering Num \\ Epochs}} &
  \multicolumn{1}{c|}{\textbf{$\alpha$}} &
  \multicolumn{1}{c|}{\textbf{$\beta$}} \\ \hline
\multirow{3}{*}{R\_CE} &
  1 &
  \multirow{6}{*}{2e-05} &
  \multirow{6}{*}{20} &
  \multirow{6}{*}{250} &
  3 &
  \multirow{3}{*}{N/A} &
  \multirow{3}{*}{N/A} \\ \cline{2-2} \cline{6-6}
                                & 2 &  &  &  & 5 &                    &     \\ \cline{2-2} \cline{6-6}
                                & 3 &  &  &  & 3 &                    &     \\ \cline{1-2} \cline{6-8} 
\multirow{3}{*}{R\_KLNLL} & 1 &  &  &  & 3 & \multirow{3}{*}{1} & 0.5 \\ \cline{2-2} \cline{6-6} \cline{8-8} 
                                & 2 &  &  &  & 2 &                    & 0.2 \\ \cline{2-2} \cline{6-6} \cline{8-8} 
                                & 3 &  &  &  & 2 &                    & 0.2 \\ \hline
\end{tabular}
\label{tabparam}
\end{center}
\end{table}

\subsubsection{\textbf{Multitask Training}} In contrast, in multitask setting, a single model is trained to produce outputs for all 3 tasks simultaneously. As illustrated in Fig. \ref{fig:modelarch_b}  all components up to and including the intermediate linear layer are shared, with only the final classification layers separate for each task. The final loss function is now a weighted combination of all 3 tasks, as shown in (\ref{eqmulti}) where $\lambda\textsubscript{t}$ and $\mathcal{L}\textsubscript{t}$, denotes the weight and value of each component's loss, respectively. 

\begin{equation} 
\mathcal{L} = \sum_{t=1}^{3}\lambda\textsubscript{t}\mathcal{L}\textsubscript{t}
\label{eqmulti}
\end{equation}

$\lambda$'s are tuned among candidate values (0.2, 0.5, 0.8, 1) also using 4-fold cross validation on the train set. Values that yield the highest average macro-F1 score for all 3 tasks are selected for final training. Multitasks RoBERTa\_CE models are trained for 3 epochs while its KL\_NLL counterpart only 2. To achieve reported performance, Multitask RoBERTa\_KLNLL's learning rate of Task 1's classification layer is set to 2.5e-5, and  final $\lambda$'s to 0.2, 0.2, 1 for Task 1, 2 and 3, respectively. All other hyperparameters stay consistent with their reported counterparts in Table \ref{tabparam}.

\begin{table*}[htbp]
 \caption{Macro results for Single and Multitask Settings.}
\centering
\begin{adjustbox}{max width=\textwidth}
    \begin{tabular}{|lccccccccc|}
\hline
\multicolumn{10}{|c|}{\textbf{Single Task Training}} \\ \hline
\multicolumn{1}{|c|}{\multirow{2}{*}{\textbf{Model}}} &
  \multicolumn{3}{c|}{\textbf{Task 1: Aggression}} &
  \multicolumn{3}{c|}{\textbf{Task 2:    Target}} &
  \multicolumn{3}{c|}{\textbf{Task 3: Label}} \\ \cline{2-10} 
\multicolumn{1}{|c|}{} &
  \multicolumn{1}{c|}{\textbf{Precision}} &
  \multicolumn{1}{c|}{\textbf{Recall}} &
  \multicolumn{1}{c|}{\textbf{F1}} &
  \multicolumn{1}{c|}{\textbf{Precision}} &
  \multicolumn{1}{c|}{\textbf{Recall}} &
  \multicolumn{1}{c|}{\textbf{F1}} &
  \multicolumn{1}{c|}{\textbf{Precision}} &
  \multicolumn{1}{c|}{\textbf{Recall}} &
  \textbf{F1} \\ \hline
\multicolumn{1}{|l|}{RoBERTa\_CE} &
  \multicolumn{1}{c|}{0.63 +- 0.04} &
  \multicolumn{1}{c|}{0.63 +- 0.02} &
  \multicolumn{1}{c|}{0.63 +- 0.03} &
  \multicolumn{1}{c|}{0.81 +- 0.01} &
  \multicolumn{1}{c|}{0.87 +- 0.02} &
  \multicolumn{1}{c|}{0.83 +- 0.01} &
  \multicolumn{1}{c|}{0.57 +- 0.03} &
  \multicolumn{1}{c|}{0.61 +- 0.03} &
  0.57 +- 0.03 \\ \hline
\multicolumn{1}{|l|}{RoBERTa\_KLNLL\textsubscript{no softmax}} &
  \multicolumn{1}{c|}{0.67 +- 0.03} &
  \multicolumn{1}{c|}{0.63 +- 0.02} &
  \multicolumn{1}{c|}{0.64 +- 0.02} &
  \multicolumn{1}{c|}{0.80 +- 0.01} &
  \multicolumn{1}{c|}{0.89 +- 0.01} &
  \multicolumn{1}{c|}{0.83 +- 0.01} &
  \multicolumn{1}{c|}{0.59 +- 0.02} &
  \multicolumn{1}{c|}{0.61 +- 0.05} &
  0.59 +- 0.04 \\ \hline
\multicolumn{1}{|l|}{RoBERTa\_KLNLL\textsubscript{softmax}} &
  \multicolumn{1}{c|}{\textbf{0.72} +- 0.09} &
  \multicolumn{1}{c|}{\textbf{0.64} +- 0.01} &
  \multicolumn{1}{c|}{\textbf{0.66} +- 0.02} &
  \multicolumn{1}{c|}{\textbf{0.81} +- 0.01} &
  \multicolumn{1}{c|}{\textbf{0.89} +- 0.01} &
  \multicolumn{1}{c|}{\textbf{0.84} +- 0.01} &
  \multicolumn{1}{c|}{\textbf{0.62} +- 0.02} &
  \multicolumn{1}{c|}{\textbf{0.64} +- 0.02} &
  \textbf{0.62} +- 0.02 \\ \hline
\multicolumn{10}{|c|}{\textbf{Multitask Training}} \\ \hline
\multicolumn{1}{|l|}{RoBERTa\_CE\textsubscript{equal}} &
  \multicolumn{1}{c|}{0.60 +- 0.01} &
  \multicolumn{1}{c|}{0.66 +- 0.03} &
  \multicolumn{1}{c|}{0.62 +- 0.01} &
  \multicolumn{1}{c|}{0.80 +- 0.01} &
  \multicolumn{1}{c|}{\textbf{0.91} +- 0.01} &
  \multicolumn{1}{c|}{0.83 +- 0.00} &
  \multicolumn{1}{c|}{0.57 +- 0.02} &
  \multicolumn{1}{c|}{0.66 +- 0.03} &
  0.59 +- 0.02 \\ \hline
\multicolumn{1}{|l|}{RoBERTa\_KLNLL\textsubscript{equal}} &
  \multicolumn{1}{c|}{0.66 +- 0.03} &
  \multicolumn{1}{c|}{0.66 +- 0.02} &
  \multicolumn{1}{c|}{0.66 +- 0.02} &
  \multicolumn{1}{c|}{0.80 +- 0.01} &
  \multicolumn{1}{c|}{0.89 +- 0.01} &
  \multicolumn{1}{c|}{0.84 +- 0.01} &
  \multicolumn{1}{c|}{0.61 +- 0.01} &
  \multicolumn{1}{c|}{0.63 +- 0.02} &
  0.58 +- 0.01 \\ \hline
\multicolumn{1}{|l|}{RoBERTa\_KLNLL\textsubscript{tuned}} &
  \multicolumn{1}{c|}{0.66 +- 0.01} &
  \multicolumn{1}{c|}{0.66 +- 0.04} &
  \multicolumn{1}{c|}{0.65 +- 0.02} &
  \multicolumn{1}{c|}{0.82 +- 0.01} &
  \multicolumn{1}{c|}{0.86 +- 0.01} &
  \multicolumn{1}{c|}{0.84 +- 0.01} &
  \multicolumn{1}{c|}{0.63 +- 0.02} &
  \multicolumn{1}{c|}{0.64 +- 0.02} &
  0.62 +- 0.02 \\ \hline
\multicolumn{1}{|l|}{RoBERTa\_KLNLL\textsubscript{ensemble5}} &
  \multicolumn{1}{l|}{\textbf{0.69}} &
  \multicolumn{1}{l|}{\textbf{0.66}} &
  \multicolumn{1}{l|}{\textbf{0.67}} &
  \multicolumn{1}{l|}{\textbf{0.83}} &
  \multicolumn{1}{l|}{0.86} &
  \multicolumn{1}{l|}{\textbf{0.84}} &
  \multicolumn{1}{l|}{\textbf{0.65}} &
  \multicolumn{1}{l|}{\textbf{0.67}} &
  \multicolumn{1}{l|}{\textbf{0.65}} \\ \hline
\end{tabular}
\end{adjustbox}
    \label{tabresults}
\end{table*} 


\begin{table*}[htbp]
\caption{Results by category of RoBERTA\_CE\textsubscript{ensemble5} \textit{(left sub-columns)} and RoBERTA\_KLNLL\textsubscript{ensemble5} \textit{(right sub-columns)} .}
\centering
\begin{tabular}{l|cccccc|cccc|cccccc|}
\cline{2-17}
 &
  \multicolumn{6}{c|}{\textbf{Task 1}} &
  \multicolumn{4}{c|}{\textbf{Task 2}} &
  \multicolumn{6}{c|}{\textbf{Task 3}} \\ \cline{2-17} 
 &
  \multicolumn{2}{c|}{\textit{Not}} &
  \multicolumn{2}{c|}{\textit{Somewhat}} &
  \multicolumn{2}{c|}{\textit{Very}} &
  \multicolumn{2}{c|}{\textit{N/A}} &
  \multicolumn{2}{c|}{\textit{anti-Asian}} &
  \multicolumn{2}{c|}{\textit{Normal}} &
  \multicolumn{2}{c|}{\textit{Abusive}} &
  \multicolumn{2}{c|}{\textit{Hate}} \\ \hline
\multicolumn{1}{|l|}{\textbf{Precision}} &
  \multicolumn{1}{c|}{0.91} &
  \multicolumn{1}{c|}{0.90} &
  \multicolumn{1}{c|}{0.61} &
  \multicolumn{1}{c|}{0.69} &
  \multicolumn{1}{c|}{0.28} &
  0.47 &
  \multicolumn{1}{c|}{0.99} &
  \multicolumn{1}{c|}{0.96} &
  \multicolumn{1}{c|}{0.61} &
  0.69 &
  \multicolumn{1}{c|}{0.95} &
  \multicolumn{1}{c|}{0.93} &
  \multicolumn{1}{c|}{0.38} &
  \multicolumn{1}{c|}{0.54} &
  \multicolumn{1}{c|}{0.39} &
  0.47 \\
\multicolumn{1}{|l|}{\textbf{Recall}} &
  \multicolumn{1}{c|}{0.89} &
  \multicolumn{1}{c|}{0.93} &
  \multicolumn{1}{c|}{0.61} &
  \multicolumn{1}{c|}{0.62} &
  \multicolumn{1}{c|}{0.50} &
  0.44 &
  \multicolumn{1}{c|}{0.89} &
  \multicolumn{1}{c|}{0.94} &
  \multicolumn{1}{c|}{0.94} &
  0.79 &
  \multicolumn{1}{c|}{0.87} &
  \multicolumn{1}{c|}{0.93} &
  \multicolumn{1}{c|}{0.43} &
  \multicolumn{1}{c|}{0.45} &
  \multicolumn{1}{c|}{0.70} &
  0.65 \\
\multicolumn{1}{|l|}{\textbf{F1}} &
  \multicolumn{1}{c|}{0.90} &
  \multicolumn{1}{c|}{0.91} &
  \multicolumn{1}{c|}{0.61} &
  \multicolumn{1}{c|}{0.65} &
  \multicolumn{1}{c|}{0.36} &
  0.46 &
  \multicolumn{1}{c|}{0.94} &
  \multicolumn{1}{c|}{0.95} &
  \multicolumn{1}{c|}{0.74} &
  0.74 &
  \multicolumn{1}{c|}{0.91} &
  \multicolumn{1}{c|}{0.93} &
  \multicolumn{1}{c|}{0.40} &
  \multicolumn{1}{c|}{0.49} &
  \multicolumn{1}{c|}{0.50} &
  0.50 \\ \hline
\end{tabular}
\label{tab:ensemble}
\end{table*}

\section{Results}
Due to the small size of our dataset, each model is trained and evaluated over 5 random seeds. In Table \ref{tabresults}, we present the average and standard deviation (SD) of macro Precision, Recall and F1 scores over these runs to compare the models' performance in each setting.

\subsection{Single Task Performance} 
We observe that RoBERTa\_CE provides respectable performance. Unsurprisingly, the model achieves the highest F1 score at above 0.8 on the binary classification Task 2 of predicting targets. Task 1 yields the next best result with F1 reaching 0.63 while Task 3 emerges the most challenging, yielding F1 not reaching 0.6. Overall, this discrepancy among results appear consistent with the corresponding attribute's level of (dis)agreement among annotations discussed in Section III.

For RoBERTa\_KLNLL, we also present results where the raw soft labels $p(y)$ are use instead of $p\textsubscript{softmax}(y)$ to calculate the KL Divergence component in \eqref{eqklnll} as a reference point. This RoBERTa\_KLNLL\textsubscript{no\_softmax} model does achieve 4 points and 1 point respective increase in Task 1's Precision and F1 scores, and 2 points for Task 3's. These considerable improvements demonstrate the efficacy of leveraging soft labels' information to assist the model in predicting the final silver labels. By minimizing the KL divergence from the soft labels, the model is equipped to predict distribution of annotations, enhancing the ultimate classifications. 

Using $p\textsubscript{softmax}(y)$ allows RoBERTa\_KLNLL (denoted RoBERTa\_KLNLL\textsubscript{softmax}) to gain further improvements across the board. In fact, Task 2 now sees a 1 point increase in F1 (0.84) over the other 2 models whereas Task 3's F1 score now jumps to 0.62 from RoBERTa\_CE's 0.57, an 8.8\% boost. We acknowledge that this model's Precision score of 0.72 may be inflated by some outlier results (0.9 SD). However, the corresponding F1 score of 0.66 corroborates RoBERTa \textsubscript{softmax}'s superior performance.\\

\noindent{\textbf{Why $p\textsubscript{softmax}$?}} Applying the \textit{Softmax} function to raw soft labels, which are already discrete probability distributions themselves, reduces the overall range of the distribution. For example, a soft label distribution of $[1.0, 0., 0.]$ then becomes approximately $[0.6, 0.2, 0.2]$ post-\textit{Softmax}, with the index of the maximum value unchanged. On the other hand, an equally divisive tweet of \textit{Annotation Class C} retains its soft label of $[0.\bar{3}, 0.\bar{3}, 0.\bar{3}]$ regardless. RoBERTa\_KLNLL\textsubscript{softmax}'s most significant improvement in F1 scores over RoBERTa\_CE in Task 3, which has the most diverse \textit{Annotation Class} distribution among all tasks, demonstrates the advantage of minimizing the KL Divergence from the flattened $p\textsubscript{softmax}$ distribution. Furthermore, the $NLL(\;p(\hat{y}\textsubscript{correct}))$ component is expected to guide the model to allocate more probability mass towards the correct class, especially in cases of ambiguous agreement.
\subsection{Multitask Performance} 
In Table \ref{tabresults}, we present the macro results for the multitask RoBERTa\_CE where $\lambda\textsubscript{i}$'s = 1, denoted RoBERTa\_CE\textsubscript{equal}, as baseline for this setting. This model yields significantly higher Recalls at the cost of generally lower Precisions, resulting in overall worse F1 scores compared to its single task counterpart. In contrast, RoBERTa\_KLNLL with $\lambda\textsubscript{i}$'s = 1, also denoted with the subscript ``\textit{equal}", achieves much better performance, with Task 1 and 2's F1 scores comparable to its single task counterparts. However, RoBERTa\_KLNLL\textsubscript{equal} under-performs in Task 3, with F1 score of only 0.58. 

Using the values of {0.2, 0.2, 1} for $\lambda\textsubscript{1}, \lambda\textsubscript{2}, \lambda\textsubscript{3}$ respectively, RoBERTa\_KLNLL\textsubscript{tuned} recovers its expected Task 3's F1 score of 0.62 while only having a slightly lower F1 score in Task 1 (0.65 vs 0.66). For Task 2, RoBERTa\_KLNLL\textsubscript{tuned} yields a Recall of 0.86, 3 points lower than its single task's 0.89, but makes up with Precision of 0.82 and an F1 score equal to its single task version (0.84). We note that this set of $\lambda\textsubscript{i}$'s essentially treats Task 3 as the main task while the other 2 as auxiliary. 

\noindent{\textbf{Multitask over Single Task?}} By sharing all but the final classification layers, multitask training reduces the number of trainable parameters by almost 3 times as opposed to training separately. Furthermore, our annotation template does not allow a tweet to be labelled \textit{Not Aggressive} yet have a non-neutral Target or Speech Type. Multitask training allows RoBERTa\_KLNLL to learn this constraint, producing only 1 instance of this type of self-contradictory prediction across all 5 seeds whereas single task training produces 61 instances.

   

\noindent{\textbf{Ensemble Classifier}}  We construct an ensemble classifier, denoted RoBERTA\_KLNLL\textsubscript{ensemble}, by combining the results of each independent seed's run via majority voting. The final predicted label per attribute for each tweet is the one with the highest frequency among all seeds' corresponding votes. Ties are decided using the following simple logic:

\begin{itemize}
    \item \textit{Aggression}: \textit{(Not, Somewhat) $\rightarrow$ Not, (Somewhat, Very) $\rightarrow$ Somewhat, (Not, Very) $\rightarrow$ Not  }
    \item \textit{Target: (N/A, anti-Asian) $\rightarrow$ N/A}
    \item \textit{Type: (Normal, Abusive) $\rightarrow$ Normal, (Abusive, Hate) $\rightarrow$ Abusive, (Normal, Hate) $\rightarrow$ Normal }
\end{itemize}

\begin{figure}[htbp]
    \centering
    \includegraphics[width=\columnwidth]{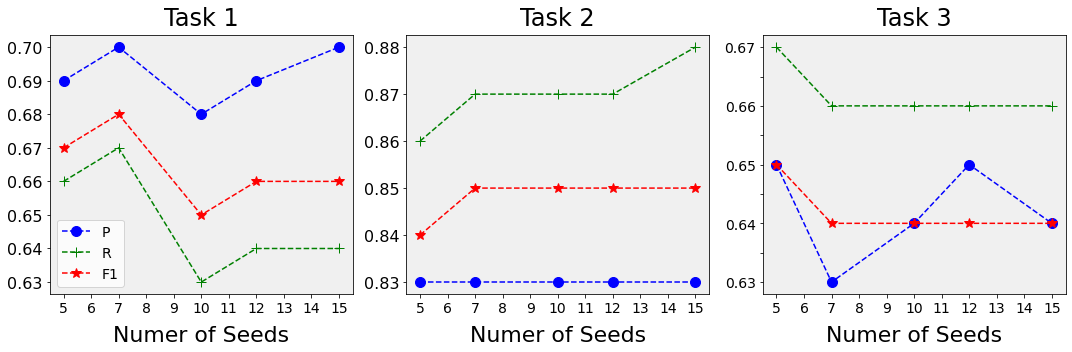}
    \caption{Macro Results of RoBERTA\_KLNLL\textsubscript{ensemble} using different number of seeds.}
    \label{fig:ensemble}
\end{figure}

Fig. \ref{fig:ensemble} displays the macro Precision, Recall and F1 scores over the test set for each task using different number of seeds. We report the set of results using 5 seeds in Table \ref{tabresults} as RoBERTa\_KLNLL\textsubscript{ensemble5} for consistency. Task 1 and 2 achieve peak macro F1 scores at 7 seeds (0.68 and 0.85, respectively) while Task 3 at only 5 seeds (0.65), allowing better computational efficiency over using more (10+) seeds.
Furthermore, RoBERTa\_KLNLL\textsubscript{ensemble5}'s results match or exceed other models' reported means in almost all metrics, validating this simple ensemble approach's effectiveness when only a single predicted label per attribute is demanded. 

Finally, Table \ref{tab:ensemble} provides a side-by-side comparison between the results per category between RoBERTa\_KLNLL\textsubscript{ensemble5} (right sub-columns) and the similarly constructed RoBERTa\_CE\textsubscript{ensemble5} (left sub-columns). The latter model tends to over-classify samples into the more minor categories (by silver label's frequency) to achieve higher Recall (Task 1 \textit{Very Aggressive}, Task 2 \textit{anti-Asian}, Task 3 \textit{Hate}) at the cost of drastically lower Precision, and consequently under-classify the major neutral categories. In contrast, RoBERTa\_KLNLL\textsubscript{ensemble5} yields much more balanced metrics both within and across categories, offering further evidence to support our method of combining soft and silver labels to combat class imbalance.

\section{Conclusion \& Future Work}
In this work, we have made 3 contributions with generalizable insights into the field of hate speech detection. First, we provide a COVID-related dataset with a focus on Asians, a group often underrepresented in literature \cite{fortuna2018survey, vidgen2020directions}. Our crowdsourced annotation is an alternative approach to expert adjudication, which provides a case study where public sentiments are attached to a still-developing social phenomenon. 

 Due to the subjectivity of personal interpretation, there are often scenarios with no clear consensus among annotators, and we argue that this information should be incorporated in downstream classification models. Our second contribution is the proposed approach to leverage soft labels, which conveys annotators' levels of agreement, via a simple modification of the objective functions. Our experiments show that this approach could produce considerably better results than using solely majority-voted singular labels. Third, we demonstrate a straightforward architecture along with a training regimen that is not only computationally parsimonious, but also adaptable to other datasets, tasks and/or more sophisticated models.

Our results corroborate the benefits of releasing individual annotator's labels in addition to the singular adjudicated label \cite{vidgen2020directions, poletto2021resources, fornaciari2021beyond, davani2022dealing}. We acknowledge the limited size of our dataset, which is a result of resource constraints, a common challenge in procuring data for under-represented groups \cite{vidgen2019challenges}. Furthermore, ``Asians" and ``Blacks" consist of many subgroups with diverse characteristics. We invite other researchers to extend their work towards these subgroups beyond the general umbrella categories introduced in this paper. Finally, the vast portion of available data is often unlabelled. We believe further investigation into meta learning or semi-supervised learning approaches should be considered to more effectively leverage this generally under-utilized resource.

\section{Acknowledgement}

\section*{Appendix}\label{apdx:phrases}
\small
\noindent{Due to space constraints, anti-Black key phrases will be released with the official repository.}\\
\noindent{\textbf{Anti-Asian Phrases}} batsoup,  bioattack,  blame  china,  boycott  china,  bug  men, bugland,  ccp,  chankoro,  chicom,  china  is  asshole,  chinais  terrorist,  china  lie  people  die,  china  should  apologize, china  virus,  china  virus  outbreak,  chinaflu  ,  chinazi,  chi-nese  propaganda,  chinese  virus,  ching  chong,  chinigger, chink,  chinkland  ,  chinksect,  communism  kill,  communistchina, fuck china, goloid, gook, gook eyed, gookie, gook-let, gooky eye, insectoid, make china pay, no asian allowed, no  chinese  allowed,  oriental  devil,  pinkdick,  ricenigger, wohan,  wuflu,  wuhancorona,  wuhaninfluenza,  wuhanpne-unomia, wuhansars, yellow jew, yellow nigger, yellow peril

\bibliographystyle{IEEEtran}
\bibliography{mainbib}

\end{document}